\documentclass[eng]{IEEEtran}
\pdfoutput=1 
\usepackage[english]{babel}
\usepackage{amsmath}
\usepackage{graphicx}
\usepackage{booktabs}
\usepackage[labelfont=bf]{caption}
\usepackage{caption}
\usepackage{subcaption}
\usepackage{hyperref}
\usepackage{xcolor}

\title{Training neural networks with synthetic electrocardiograms}

\author{\IEEEauthorblockN{Matti Kaisti$^1$\IEEEauthorrefmark{1},
Juho Laitala$^1$, and
Antti Airola$^1$}

\IEEEauthorblockA{$^1$ Department of Computing, Digital Health Lab, University of Turku, Turku 20500, Finland}
\thanks{\IEEEauthorrefmark{1}Corresponding author:mkaist@utu.fi }}

\begin{document}
\maketitle

\begin{abstract}
We present a method for training neural networks with synthetic electrocardiograms that mimic signals produced by a wearable single lead electrocardiogram monitor. We use domain randomization where the synthetic signal properties such as the waveform shape, RR-intervals and noise are varied for every training example. Models trained with synthetic data are compared to their counterparts trained with real data. Detection of r-waves in electrocardiograms recorded during different physical activities and in atrial fibrillation is used to compare the models. By allowing the randomization to increase beyond what is typically observed in the real-world data the performance is on par or superseding the performance of networks trained with real data. Experiments show robust performance with different seeds and training examples on different test sets without any test set specific tuning. The method makes possible to train neural networks using practically free-to-collect data with accurate labels without the need for manual annotations and it opens up the possibility of extending the use of synthetic data on cardiac disease classification when disease specific a priori information is used in the electrocardiogram generation. Additionally the distribution of data can be controlled eliminating class imbalances that are typically observed in health related data and additionally the generated data is inherently private. 
\end{abstract}


\section{Introduction}

Training neural networks typically requires significant amount of labeled data that are expensive to collect and this is especially true for healthcare data where expert knowledge is required~\cite{bote2019deep} and open sharing is limited due to privacy concerns~\cite{rieke2020future}. Typically, better predictive performance in deep learning is sought after by using more data and/or more complex and bigger networks~\cite{muralidhar2018incorporating}. However, in the field of healthcare, this approach has several shortcoming as i) medical healthcare data is expensive to gather, ii) labeling can be difficult due to unspecificity and co-diseases, iii) data classes are typically unbalanced, iv) poor and varying signal quality and v) increasingly strict privacy legislation makes obtaining and sharing real-world patient data challenging.

An approach to address these shortcomings using free synthetic data with accurate labels has been shown to be promising in image classification tasks~\cite{tobin2017domain, tremblay2018training, prakash2019structured, james2019sim}. A clear advancement was achieved with domain randomization where photorealism, a requirement of earlier attempts, was abandoned and random perturbations of the environment in non-realistic ways achieved competitive accuracy in testing~\cite{tremblay2018training}. The network learned to discriminate between desired and undesired objects by adding randomly different geometric shapes, random textures and random lights to the images. Another line of research in health monitoring has shown the benefit of using pre-trained image nets and transfer learning for scarce 1D health data applications where the signal is first transformed into an image which is followed by a fine tuning of the model weights for final predictive model~\cite{salem2018ecg, weimann2021transfer}. This removes the need for large application specific datasets. However, 1D signals are not always well presented as images and either the morphological details or long term information ~\cite{costa2002multiscale} of the signal is invariably lost. Furthermore, the properties of networks employing memory properties~\cite{ordonez2016deep} are better suited for 1D signals and pre-trained networks are bounded by the approach chosen during initial training which could be unoptimal for the task at hand.   

\begin{figure*}
\centering
\includegraphics[width=\textwidth]{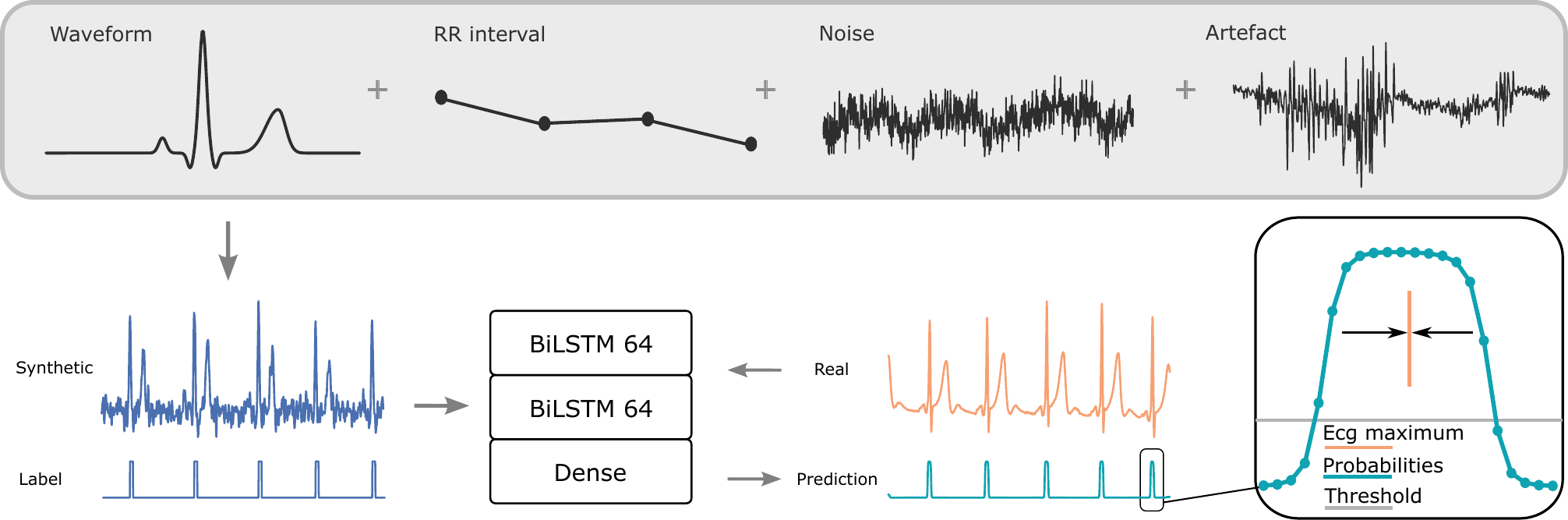}
\caption{\label{fig:Fig1} Principle of synthetic data generation, model training and testing and post-processing step. Mathematical model is used to generate synthetic data where the properties of the waveform shape, RR-intervals and noise can be controlled. Real artefacts are optionally extracted, randomized and added to the synthetic signals. A label array with a same length than the corresponding ECG are used and this array has five one's at the r-wave location and zero's elsewhere. The prediction probabilities are transformed to a single peak location in the post-processing step.}
\end{figure*}

In this work we describe a synthetic signal generator that is able to produce electrocardiograms (ECGs) where characteristics of these signals can be varied in a controlled manner and in part solve and investigate the above mentioned challenges. Here, ECGs are used to train an LSTM network and we demonstrate the learning through r-wave detection using various real-life recordings in testing. The procedure of signal generation is exemplified in Figure~\ref{fig:Fig1} and comprises the generation of i) signal waveform, ii) RR-intervals, iii) noise process and iv) augmentation of real data artefact. Synthetic signals with varying degree of domain randomization are fed to the network and final estimate on detected peak indices is achieved through a post-processing step. We show that neural networks can be effectively trained and such models can achieve better results compared to models trained with real data and that the models trained with synthetic data are robust against different datasets without any input data specific hyperparameter tuning. 

\section{Methods}
\subsection{Synthetic electrocardiogram generator}
\label{section:generator}
The ECG signal generator is comprised of four main parts. i) RR-interval generation where the average interval (HR) and its variation (HRV) can be adjusted. ii) Waveform model where each of the characteristic waves (p, q, r, s, t) of an ECG can be independently adjusted in terms of amplitude (positive and negative), width and location. We refer the wave locations as fiducial points.  iii) Generalized noise model allowing a noise realization to be generated from an arbitrary spectrum allowing for example 1/f and random walk noises to be generated that roughly resemble motion artefacts and iv) augmentation of real artefacts where a random segment from recorded artefact signals is added to the generated ECG with random amplitude.

Each synthetic signal is generated using model input parameters. We allow these parameters to change within a predefined range from which we extract randomly and independently each parameter value using a uniform distribution. The limits of this range, and thus randomization and variance of the input data, is controlled using a scaling coefficient (C) which in controlled independently for RR-intervals, waveform shape, fiducial point and noise. Same C was used for each independent part if not stated otherwise. Scaling coefficient C = 1 mimics physiological variance range of the healthy. These randomization limits, that were subsequently scaled, were determined by a combination of values in literature~\cite{thaler2017only} and by fitting the model to real measurements with visual comparison. Some fitting examples are shown in Figure~\ref{fig:waveform_fittings} which are also used to validate the model. The C multiplies the range defined with lower and upper limits $l_{low}$ and $l_{high}$ of each adjustable parameter. The midpoint shifts non-linearly as weighted scaling is used as $l_{low} - d$ and $l_{high} + d$ where $d = |l_{low} - l_{high}|\times (C-1) \times l_{low}/(l_{low} + l_{high}) $. This allows parameter limits close to zero to vary more gently. However, this does not exclude inverted waves when scaling is sufficiently large. Additionally, the t-wave locations were made dependent on heart rate where the distance to r-wave further scales with square root of the average RR-interval~\cite{vandenberk2016qt}. The lower limit for all noise randomizations is zero and this does not change during scaling. This allows some degree of low noise signals to always be in the training set. The starting limits for noise are subjective as the level of noise can vary significantly between devices and situations. The noise limits are adjusted so that the r-waves in most cases are still visually separable from noise when C = 3. The starting limits are listed on Table~\ref{tab:initial ranges}. Overall the randomization procedure is subjective, but it roughly follows the principle of starting from signal variation of the healthy in rest and increasing it as high as the model allows when C = 3.       

\subsubsection{RR-intervals}
The RR-intervals are modeled as 

\begin{equation}
rr_i = \mu + \beta sin(2\pi f_b t_i) + \gamma
\end{equation}

\noindent where $\mu$ is the average RR-interval, the second term is the breathing modulation with coupling coefficient $\beta$, breathing frequency $f_b$ and $t_i$ is the sum of previous intervals and $\gamma$ presents a stochastic component including long term correlation between RR-intervals~\cite{kantelhardt2003modeling}. The last term for the presented method is not required as the training is done on short signal segments.\\

\subsubsection{Electrocardiogram waveform} The ECG waveform (p, q, r, s and t waves) are modeled using Gaussian function as the basis~\cite{mcsharry2003dynamical} for each wave,

\begin{equation}
\dot{z_i} = \frac{-2\pi ma \phi}{b^2} \exp\bigl(\frac{-m\phi^2}{2b^2}\bigr)  
\end{equation}

\noindent where $\phi$ is a phase signal (linearly increasing) with amplitudes $[-\pi, \pi]$ where each phase cycle contains $rr_i * f_s$ samples. A separate phase signal is constructed for each wave. The time difference between the waves is achieved by simply offsetting the beginning of each phase signal by a delay that corresponds the time difference of a particular event to r-wave. An asymmetry parameter m is used to create a slightly asymmetric t-wave as typically observed in healthy. Different values of m are given to  positive and negative gradient parts to create asymmetrical shape. The gradient signals of every wave are summed,

\begin{equation}
\dot{ecg} = \sum\limits_{i \in \{p,q,r,s,t\}} \dot{z_i}
\end{equation}
The final ECG is obtained with a cumulative numerical integration of $\dot{ecg}$. \\ 

\subsubsection{Noise realizations} Time domain noise realization including white and power-law noise that corresponds a given power spectrum was generated~\cite{timmer1995generating,kaisti2016radiometric}. First a power spectral density (PSD) was defined as 

\begin{equation}
PSD = \frac{\rho}{f^{\alpha}} + \sigma^2
\end{equation}

\noindent where the components are power-law (first) and white noise (second) components. The exponent $\alpha$ is used to increase the low frequency ($f$) noise and when it is e.g 1 or 2 it reduces to 1/f noise and random walk, respectively and $\rho$ is a constant. This power spectrum is converted to time domain noise realizations by first multiplying the amplitudes of each frequency bin with an independent zero-mean complex Gaussian random variable of unit variance. Then an inverse-FFT of the randomized spectrum was computed and the real part was kept.

\subsubsection{Artefact augmentation of the electrocardiograms}
We augmented ECG signal (real and synthetic) with real ECG noise sources~\cite{laitala2020robust}. In this approach, baseline wander (BW) and muscle artifact (MA) noises from MIT-BIH Noise Stress Test database~\cite{moody1984noise, goldberger2000physiobank} and a simple generated 60 Hz sine wave representing the powerline interference are added to ECG signals with varying amplitude. Artefact realizations are obtained by randomly selecting a segment of 1000 samples (same length as the ECG segment) from both BW and MA noise sources. These segments are then multiplied by random numbers from different uniform distributions to alter the strength of the noises. In the case of BW uniform distribution of [0,10] is used and for MA it is [0,5]. The augmented artefact is one of three different categories; pure BW, pure MA or a combination of these two. After noise type selection, 60 Hz sine wave representing powerline interference is added to the noise. Magnitude of unit amplitude sine wave is varied before addition by multiplying it with random number from uniform distribution of [0,0.5]. The ranges of used uniform distributions were determined visually and the generated training examples were normalized to [-1,1] range before adding the artefact to it.

\subsection{Datasets}
Four different electrocardiogram datasets were used in this work, Glasgow University ECG database (ECG-GUDB)~\cite{https://doi.org/10.5525/gla.researchdata.716}, MIT-BIH Arrhytmia database~\cite{moody2001impact, goldberger2000physiobank}, MIT-BIH Noise Stress Test database~\cite{moody1984noise, goldberger2000physiobank} and Computing in Cardiology 2017 single atrial fibrillation database (Cinc2017-AF).~\cite{clifford2017af, PhysioNet} Both MIT-BIH databases were used only for model training while ECG-GUDB and Cinc2017-AF was used solely for testing. This was done in order to test if the trained models can generalize outside their training data. All four databases are publicly available.

ECG recordings of the ECG-GUDB database were obtained from 25 different subjects while performing five different activities (walking, jogging, operating a hand bike, solving maths test and sitting). Each task was recorded with two different setups, loose cables (standard Einthoven leads I-III) and chest trap. Therefore ECG-GUDB contains a total number of 250 (25x5x2) different ECG recordings. However, only 229 ECG records have annotations available. All ECG recordings were collected with Attys Bluethooth data acquisition board that had a sampling frequency of 250 Hz. All r-wave labels were shifted to maximum within a 16 sample window to ensure accurate labeling scheme.

In this work, we use Einthoven lead II from the loose cables setup and chest strap ECG.
We split each of the 229 records into 29 separate four second segments that are not overlapping. Thus, in total we use 6641 (229x29) four second long ECG segments for testing. The heart rate distribution is shown in Figure~\ref{fig:waveform_fittings}.

MIT-BIH Arrhytmia database contains 48 half-hour ambulatory ECG records from 47 subjects. Each record has two channels and all records have been sampled at 360 Hz. These recordings were resampled to 250 Hz. All r-wave labels were shifted to maximum within a 16 sample window to ensure accurate labeling scheme. 

MIT-BIH Noise Stress Test database has three different half-hour recordings noise: baseline wander, electrode motion artifact and muscle artifact. These recordings represent noise sources typically present in the ambulatory ECG recordings. Segments of noise were collected with electrode placement where ECG is not observable. Segments with similar noise type were concatenated into single noise records.

The Cinc2017-AF database contain single lead recordings collected with AliveCor device. The training set has 8528 recordings lasting from 9 s to just over 60 s and contains normal sinus rhythm, atrial fibrillation and alternative rhythm. The signals do not have annotated peak labels and the validation was done visually by plotting detected peak onto the signal under test. From the database we randomly selected 30 measurements labeled as atrial fibrillation. These signals varied from 15 beats to 120 beats and had in total 1336 beats. The signals are recorded by placing a finger from both hands on the metal plates of the device. Such dry electrode configuration is prone to artefacts. No artefacts were removed and if an r-wave could not be reliably identified, it was not labelled as such.

\subsection{Neural network}
For all experiments a neural network consisting two bidirectional LSTM layers with 64 units with return sequences set to True followed by a dense layer with sigmoid activation was used with Tensforflow 2.6.0. Each input sample is 1000 samples long presenting 4 s ECG (sampling frequency of 250) and the output of the model is a 1000 samples long segment where each sample is probability of that sample being an r-wave. An r-wave in training data is defined as a five neighboring one's centered at the r-wave maximum. The training is done through a generator function that provides either only real or only synthetic samples and with optional real artefact augmentation. The artefact augmentation is independent on the source of ECG samples. Each training is run with a batch size of 32 and step size of 20 for 30 epochs. Binary cross entropy was used as the loss function while Adam~\cite{kingma2014adam} was used as optimizer with a learning rate of 0.0003.  

The operating principle of the generator function that constructs the training samples can be summarized in five steps i) Real data: Select randomly 1000 sample segment from randomly selected ECG recording.  Synthetic data: Generate unique random realizations in the generator function directly when needed. ii) Generate label vector for every segment based on r-wave indices. The vector has five ones at each r-wave and is zero elsewhere. iii) If artefact augmentation is used then normalize the segment to [-1,1] range and add the generated artefact. iv) Filter the signals with a simple two order Butterworth filter with corner frequencies at 0.5 and 50 Hz. v) Normalize segment to [-1,1] range. Same filtering and amplitude normalization (steps iv and v) as above was used for test signals.

\subsection{Peak detection post-processing}
LSTM model predictions are a sequence of probability values that indicate the likelihood of a sample being an r-wave and thus the unambiguous peak location needs to be evaluated from these probability vectors. We followed similar steps as presented in~\cite{laitala2020robust}: i) Split the ECG into segments of 1000 samples with 750 sample overlap. ii) Use LSTM model to predict sample wise r-wave probabilities for each segment. iii) Take the average probabilities from overlapping predictions for each sample. Because ECG segments overlapped, four predictions are produced for each time step of the ECG signal. Overlapping predictions are averaged to get a single probability value for each sample. iv) Extract r-wave locations from average probabilities by selecting averaged probability values that are above predefined threshold of 0.05. These are considered as r-wave candidates. To produce only one peak index for every r-wave each probability candidate are shifted to index where ECG has the highest amplitude within ten sample window. When five or more samples are shifted onto the same index it is considered as an r-wave. v) Filter out r-waves that occur unrealistically close. After unique index extraction there might be some false positives e.g. pronounced t-waves or noise peaks that were identified as an r-wave. The r-waves that do not have any other r-waves within threshold distance of 75 samples are considered as valid r-waves and they form the initial set of approved r-waves. All r-waves that occur within the threshold are put into own a separate candidate set. Then the candidate set is iterated over by starting from the candidate with highest probability value. In each iteration, the candidate under consideration is compared to the set of approved r-waves. If the candidate is not within the threshold distance of any of the approved r-waves it is considered to be a valid r-wave and it is added into the set of approved r-waves.

\subsection{Code and data availability}
The neural network, artefact augmentation and peak detection post-processing has been described in detail in our earlier work~\cite{laitala2020robust}. Minor modifications on training hyperparameters such as learning rate, number of epochs and steps were implemented. No modification based on test set performance on either synthetic nor real data were done. Same training scheme was used throughout the experiments and all models were trained the same amount. Real data used here are all publicly available and code for synthetic data generation and training is exemplified in \cite{kaisti2021} and correspondingly training with real data is available at \cite{laitala2020}.

\section{Results}

\begin{figure}
\centering
\includegraphics[width=0.45\textwidth]{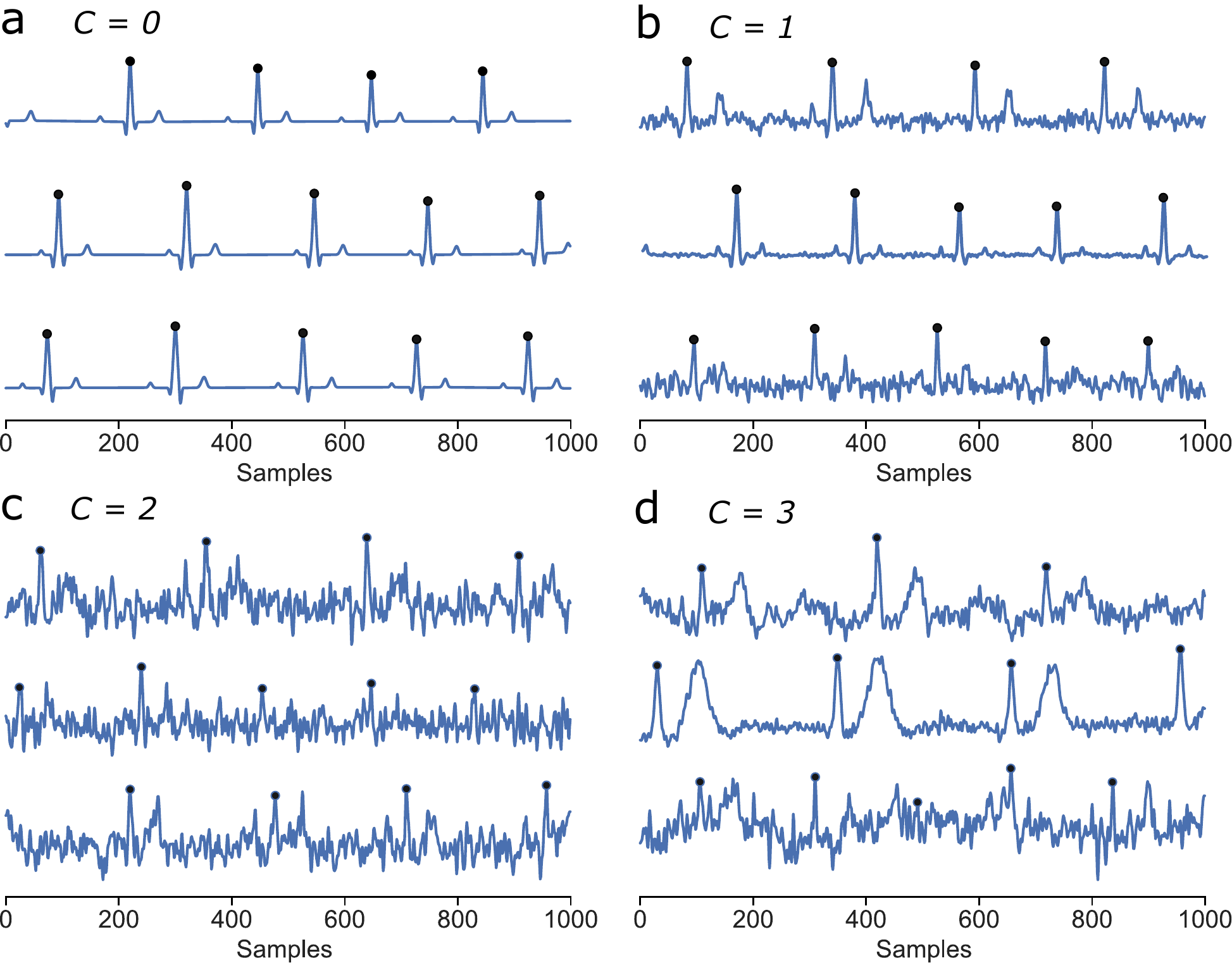}
\caption{\label{fig:Fig2} Examples of random synthetic signals generated using values of C = 0, 1, 2 and 3. C = 1 produces signals roughly within the physiological range of healthy in rest whereas C = 0, 2 and 3 produces minimal, high and very high variation between the electrocardiograms.}
\end{figure}

The variation of synthetic signals is based on the reported variation of typical ECG~\cite{thaler2017only} on the healthy and the noise properties are judged empirically from typical ECG recordings. This variation range is used as a starting point and scaled to both directions with a scaling coefficient C that signifies how much larger/smaller the range is in a particular synthetic dataset. This is detailed in Section~\ref{section:generator}. Several examples of synthetic ECGs are shown in Figure~\ref{fig:Fig2}. The similarity in a) is evident as the signals are not allowed to have any variance. However, the starting point of every realization is a randomized. In b) the noise clearly distorts the waveforms and signals resemble typical lowish quality ECGs. Domain randomization in c) and d) is high producing unrealistic and noisy ECGs. 

\begin{figure*}
\centering
\includegraphics[width=\textwidth]{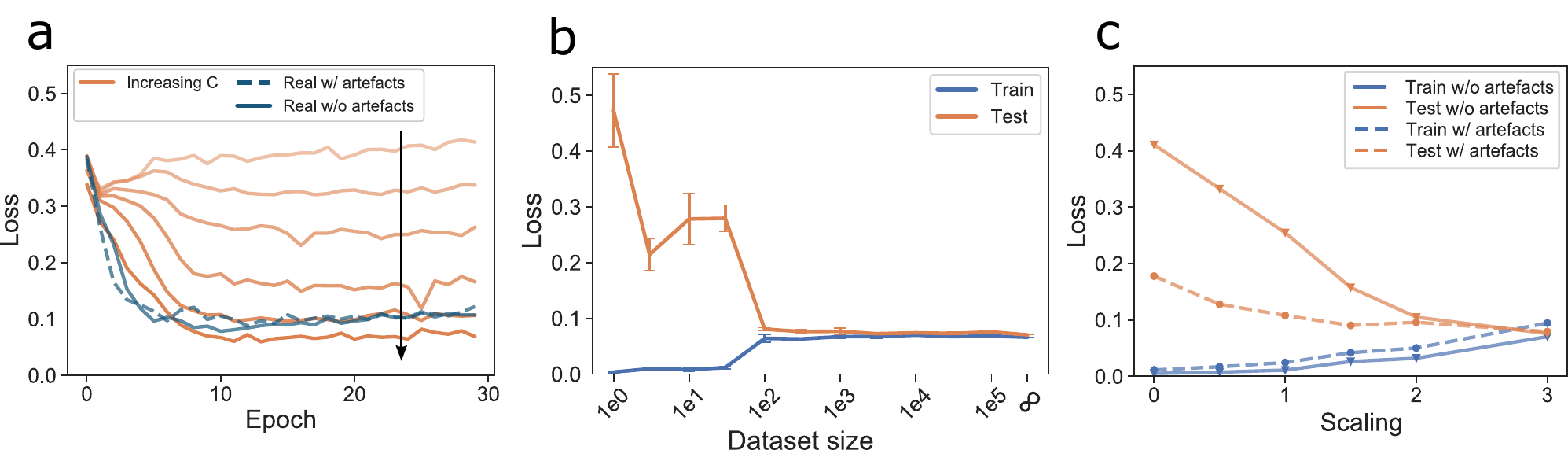}
\caption{a) Test loss during training with increasing randomization scaling coefficient, C = 0, 0.5, 1.0, 1.5, 2.0 and 3.0 (orange). Respective curves for real data with (dashed dark blue) and without (dark blue) artefact augmentation. The loss is computed after every epoch on test data. Training with synthetic data is done without artefact augmentation. b) Learning curves with C = 3 showing model performance with different input data sizes. Error bars present mean $\pm$std of repeated training's (n = 3). The $\infty$ means that every synthetic sample is unique and is generated on the fly during training. c) Training loss compared to test loss with increasing randomization. In b) and c) the test loss is an average from last five epochs.}
\label{fig:Fig3}
\end{figure*}

\begin{figure}
\centering
\includegraphics[width=0.35\textwidth]{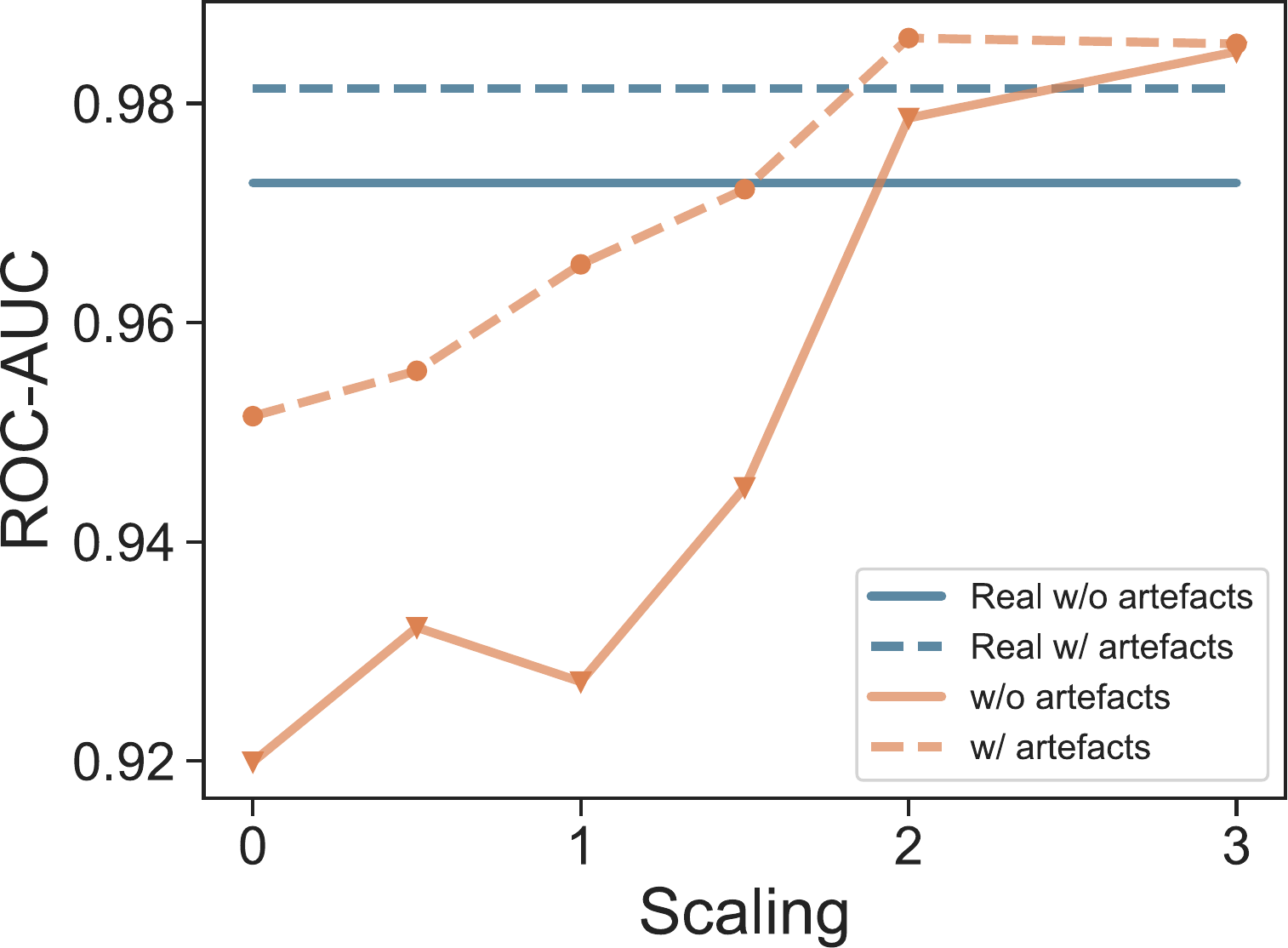}
\caption{ROC-AUC on test data (average from last five epochs) with increasing randomization scaling coefficient (C). Training with synthetic data is done without artefact augmentation.}
\label{fig:Fig3b}
\end{figure}

The loss (binary cross-entropy) computed after every epoch during the training process is shown in Figure~\ref{fig:Fig3}. In a) the test loss remains high when randomization is low and the network is unable to learn the relevant signal characteristics. The performance improves systematically as domain randomization is increased through the scaling coefficient (C) where the best performance is achieved with highest randomization when C equals 3. The testing ROC-AUC used to rank model performance in Figure~\ref{fig:Fig3b} also improves with increasing C and clear performance boost is observed when C exceeds about 1.5. With low randomization the loss is getting lower while the ROC-AUC is not improving. This implies that the network is becoming more confident and its predictions are more clear, but this does not translate on better ROC-AUC. It is also likely that the testing data has fair amount of typical and high quality signals that are fairly easy to interpret, especially the negative samples are non challenging which constitute the large majority class. The harder signals, in turn, are correctly interpreted by the network only when sufficiently high randomization at the input data is present. 

It is also noteworthy that having a physiologically valid input variation does not result in a well performing model and it is required that the randomization is increased clearly beyond what is expected to be in the test data for the best performing model. Comparisons done to models trained with real data surprisingly perform worse than a model trained purely on synthetic data. This is evident in both loss and ROC-AUC where model trained with real data performs roughly equally good as model trained with scaling coefficient of 2. However, the test set and training set are different and in part most likely have non overlapping characteristics. Having larger real dataset with more variation in training is expected to improve the performance and generalization to unseen datasets. 

The learning curves are shown in Figure~\ref{fig:Fig3} b). In this experiment synthetic datasets of varying size are pre-generated and during training the examples are drawn randomly from this set. The infinite size signify that every training example is unique as in other experiments. Expectedly the training loss is very low and test loss is high when the number of input samples is minimal. The network simply learns the properties of this data and fails to generalize. As the input data size is increased the gap between training and test curves is closing and with sufficient input data size there is very small variance and bias. The learning curves are run three times with different random signal generations and the low amount of input data produces non-robust training. As the data size is sufficient the model performances converge.

The training performance with increasing scaling is examined in Figure~\ref{fig:Fig3} c) including synthetic data without augmentation as in previous experiments as well as with the artefact augmentation. We can observe that the augmentation which adds a significant amount of randomization helps the network to learn with lesser amount of randomization. When high randomization is combined with artefacts the training performance drops below that of testing indicating the high degree of challenge the network has with this training data. The best result is achieved with highest randomization scaling and the performance is comparable with the artefact augmentation. Augmentation could have higher impact if testing data was even more challenging or more corrupted by artefacts.

\begin{figure}[h]
\centering
\includegraphics[width=0.45\textwidth]{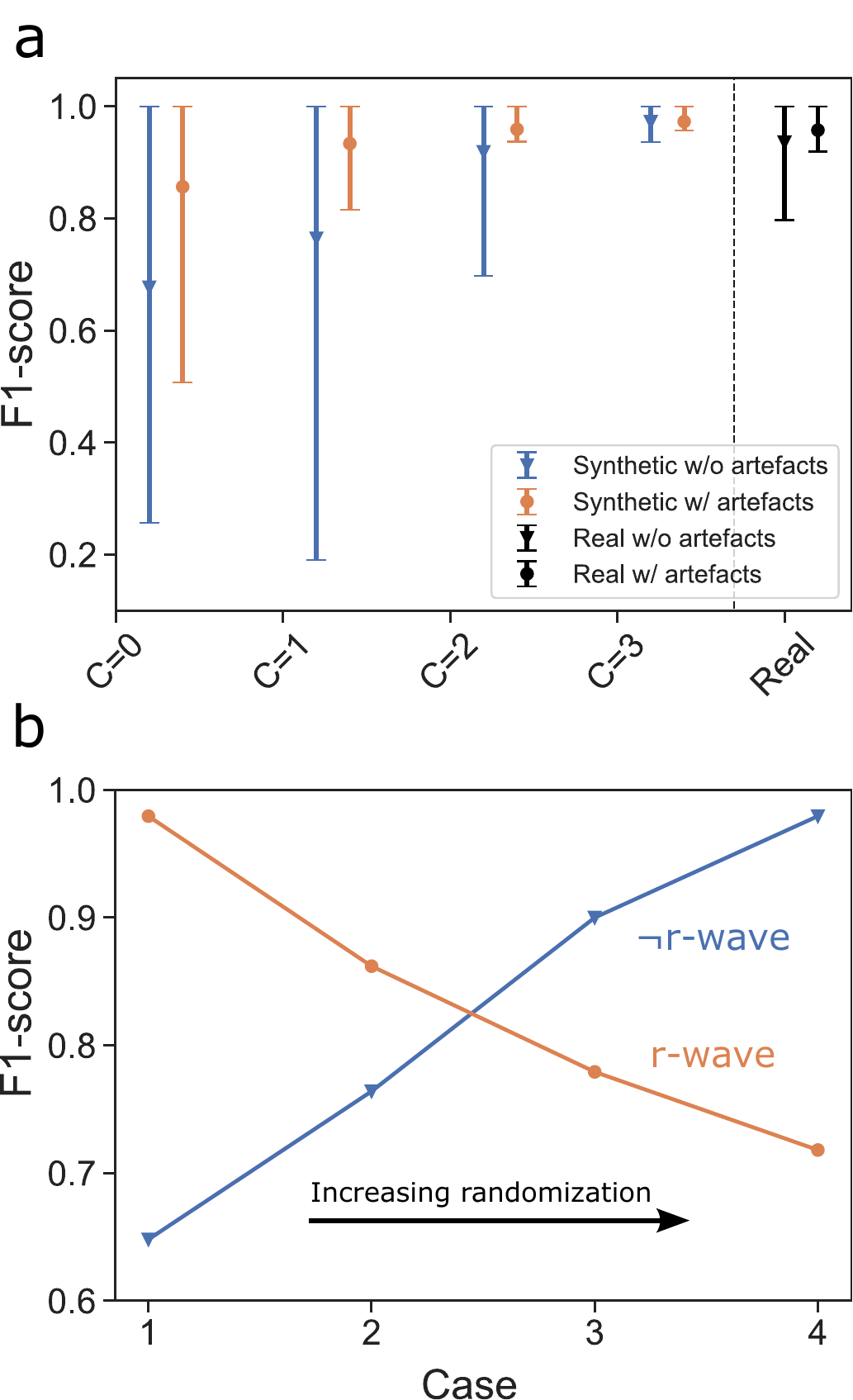}
\caption{\label{fig:Fig4} a) F1-scores on test data with scaling coefficient C = 0, 1, 2 and 3. Error bars are mean with 10\% and 90\% percentiles for F1-scores over the test data where models are trained without artefact augmentation (blue) and with augmentation (orange). Networks trained with real data are shown on the right (black). b) Performance comparison when randomization of the feature of interest (r-wave) is increased (orange) as opposed to randomizing the surroundings i.e. everything else, but the r-wave (blue). }
\end{figure}

To further validate the performance of training with synthetic data we compared the models by considering the correctly predicted peaks. Probability vectors were converted into location indices using a post-processing step. The F1-score of each measurement was used for comparison as shown in Figure~\ref{fig:Fig4} a). An F1-score was computed for each measurement. The error plots show the mean of these scores over all the measurements and whiskers present the 10\% and 90\% percentiles. The mean values systematically increase with domain randomization (increasing C) while also reducing the amount of signals where the model fails to provide meaningful result. The results also show that augmenting real artefacts onto the signals helps the model to learn which was also observed in testing loss and ROC-AUC evaluation. However, it is noteworthy that artefact augmentation alone does not yield good performance if synthetic data itself has no to minimal variation. Best results are achieved with highest randomization including artefact augmentation, but improvement is modest compared to only synthetic data with C = 3. Synthetic data without augmentation provides better performance than real data with artefact augmentation in these experiments. The overall performance compares favourably to state-of-art~\cite{peimankar2021dens}. 

Further experiments with increasing C either very slightly improved or worsened the results. This is not surprising since one criteria for scaling was that C = 3 produces the maximal amount of randomization for all adjustable parts of the signal and further increases lead to noise becoming too dominant where r-waves are completely lost, fiducial points leaking over the designated cardiac cycle and waveforms are breaking down. 

The randomization clearly has significant effect on the models ability to learn the desired characteristic. In previous experiments the r-wave was kept at a nominal and modest variance through all experiments and the surroundings were randomized as shown in Table~\ref{tab:initial ranges}. In Figure~\ref{fig:Fig4} b) we compare the effect of randomization of the desired characteristic, the r-wave, to randomization of everything else than the r-wave. In our case this can also be viewed as randomizing the samples that associate to label 1 (r-wave) or to label 0 (not r-wave) although this is not strictly true as noise realizations are added to the entire generated signal. The four cases shown in Figure~\ref{fig:Fig4} b) compare the mean F1-scores. The cases for the blue curve are with C = 0,1,2 and 3 (same as in a) ). For the orange curve the cases present increasing r-wave randomization. We chose to randomize the r-wave by adjusting the upper limit of amplitude and width parameter to match that of a t-wave while keeping the lower limit intact. This ensures that nominal r-waves are also present in all cases. The surroundings are randomized with C = 3 in all cases.  The first case is with nominal r-wave randomization (same case as the last case in previous) and the following cases (2,3,4) have upper limits matching corresponding t-wave upper limits with C = 1,2,3. As the r-wave shape randomization is increased the performance drops significantly since the model is now learning to detect various shapes in the electrocardiogram as r-waves.

The synthetic generator produces electrocardiograms where several characteristics (waveform shape, fiducial points, RR-intervals and noise) of the signal can be independently changed and randomized and thus the effect of each characteristic can be independently tested. As detailed in Table~\ref{tab:result-summary} the randomization of fiducial points had the lowest impact closely followed by RR-intervals. Randomization of the waveform shape results in clearly reduced loss, but not significantly improved ROC-AUC. This indicates that the model is able to learn easier signals with high confidence, but fails on the more difficult, mostly likely noisy ones. Addition of significant amount of noise results in a good performance measured by all metrics. This is not unexpected as the model learns to discriminate the characteristic r-wave from rest of the signal which is efficiently randomized with the addition of noise. However, experiments with only r-waves and noise (i.e. model is modified not to produce any other waves) results in modest performance and inclusion of qrs complex instead of r-wave only has negligible impact. It is, in fact, the presence t-wave that helps the model learn to discriminate between these two prominent waves. If t-wave is excluded the model is confused between t and r-wave when presented with real data. The simple combination of the r and t-wave and noise realizations performs surprisingly well and not much else is needed. However, best results are achieved with all randomization's including artefact augmentation.

One shortcoming of rule based detection algorithms is the performance during arrhythmia's which are an import to screen heart diseases~\cite{steinhubl2018effect}. We tested if the rhythm could be detected in a separate single lead atrial fibrillation test set~\cite{PhysioNet, clifford2017af} when the model is not specifically trained for it. Results are summarized in Table~\ref{table:afib}. We used  a model trained with synthetic data only (wo/ artefacts C = 3). Most false detections were due to noise artefacts in the signals or when a prominent downward qs-wave without a clear r-wave was present. The artefacts typically create a single false detection, but the model can completely fail with highly abnormal r-wave as seen in Figure~\ref{fig:afib}. In such a case the model trained with real data performed better most likely due to having some similar abnormal examples during training. Regardless, the model trained with purely synthetic data (with high domain randomization) performs well even with abnormal peak shapes, peak inversions and abnormal rhythm, none of which were specifically accounted for during training.   

\begin{table}[t]
\begin{minipage}[b]{\linewidth}
\centering
\caption{Detection of r-waves in atrial fibrillation (n = 30, no. peaks = 1336)}
\label{table:afib}
\begin{tabular}{ lccc}
    \toprule
    Model  & Precision & Recall & F1 \\
    C = 3 w/o artefacts  & 0.985  & 0.960 & 0.968\\
    Real w/ artefacts    & 0.979  & 0.981 & 0.979\\
    \bottomrule
    \addlinespace[1em]

   \end{tabular}
\end{minipage}\hfill
\begin{minipage}[b]{\linewidth}
\centering
\includegraphics[width=0.95\textwidth]{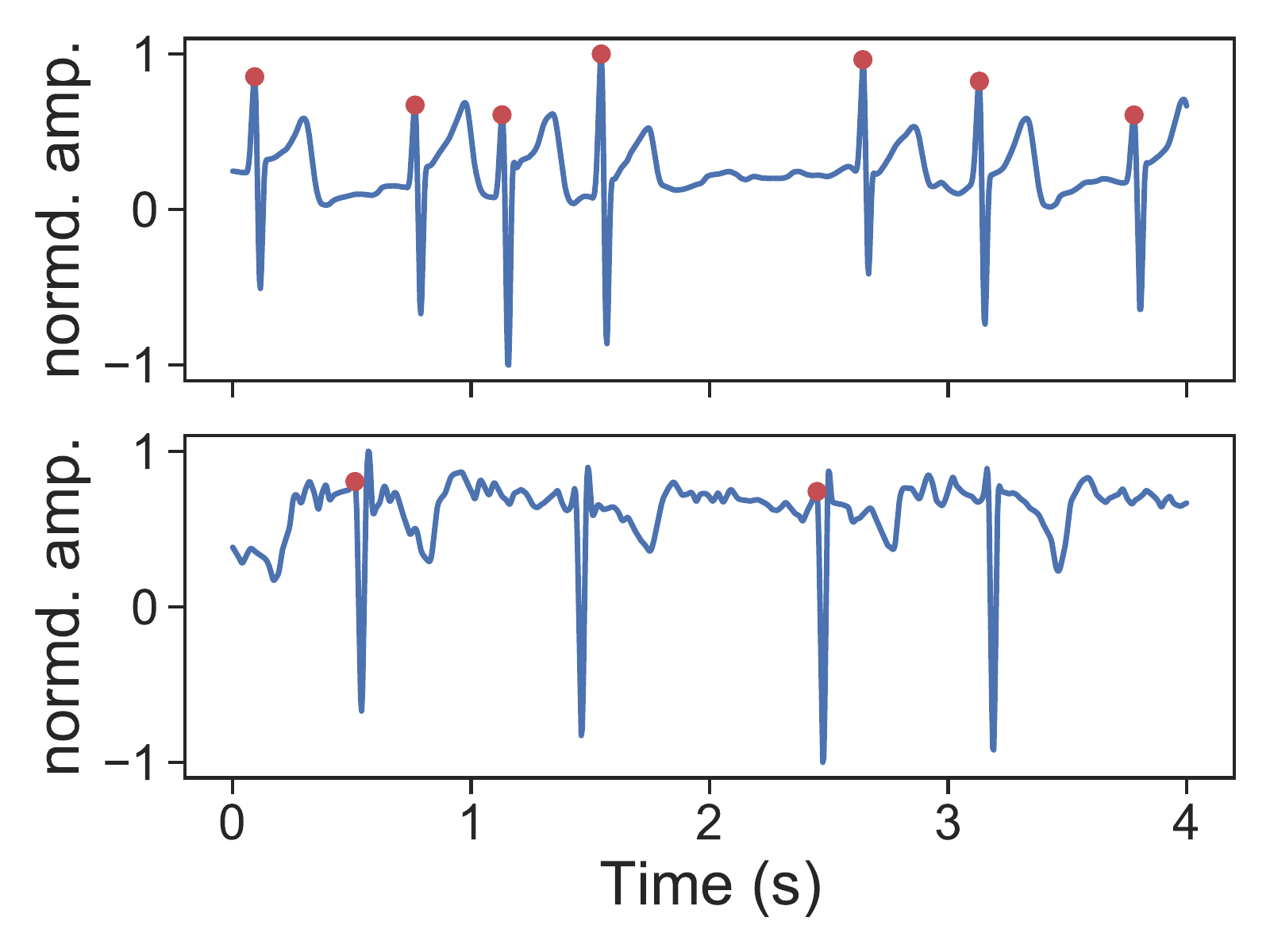}
\captionof{figure}{Examples of peak detection in atrial fibrillation with successful (upper) and unsuccessful (lower) detection. The lower graph shows a qs-wave with prominent downward deflection that the model trained with synthetic data (C=3 w/o artefacts) is unable to detect correctly.}
\label{fig:afib}
\end{minipage}
\end{table}

\section{Discussion}

The model trained with synthetic ECGs was able to not only to achieve comparable results to real data, but in-fact produce better predictive performance when tested on real data that included subjects performing various physical activities. We used r-wave detection as an example which is a prominent feature of the ECG and simply providing large enough variation provides compelling performance. Having randomization's as large as possible without drowning the r-waves in noise or breaking the model resulted in best performance. Such randomizations are clearly beyond any physiological domain or what would otherwise be expected to be present in the test signals. Further examinations with synthetic data when different randomization components (waveform, RR-intervals, noise, artefact augmentation)  are turned on individually (shown in Table~\ref{tab:result-summary}) revealed that simple noise realization is the most effective way to randomize the signal and allow the model to learn. However, if the t-wave was excluded altogether from the generated signals the performance dropped and the model is unable to discriminate between these two prominent waves. 

Our experiments also show that ECG waveform location randomizations produce smallest effect hinting that the network learns to detect the r-wave from its surroundings with little regard on what specifically happens around it and it is sufficient simply to have high variation on the parts that are not of an interest. However, the exclusion of t-wave resulted in poor performance and its presence is required as a counter example for the model. The RR-interval variation produces interesting result where the loss is high, but ROC-AUC is comparatively high highlighting the possibility that the network is able to correctly classify the labels, but with low confidence. This results in poor performance in actual r-wave detection when the post processing step is included for F1-score calculation. The insensitivity to r-wave locations was further tested in atrial fibrillation data and surprisingly the model performed well even in the presence of arrhythmia. However, with highly abnormal r-wave shapes the model occasionally failed and it does not work robustly in such special cases. This could most likely be fixed by introducing other specific r-waves in training, but this should be done in a specific and controlled manner. Overall, the model performs well against various sources of noise as exemplified in Figure~\ref{fig:predictions}. In the rare cases that a high and narrow i.e. spiky artefact is present, the model can falsely detect them as r-waves.

Additionally, the performance is robust and repeated training's with different training examples using same domain randomization scale produces closely matching results. Hyperparameter tuning and/or longer training would most likely increase the performance further. Although the synthetic data outperformed the models trained with real data, it is likely that adding significant amounts of or highly varying real data from various sources and activities would reduce or flip the performance gap.

The presented method achieved good performance on challenging electrocardiogram test sets including recordings during various physiological activities and atrial fibrillation using only one training scheme with high amount of domain randomization. This approach could be beneficial in training robust networks for various health monitoring applications and it could be extended to cardiac disease detection using vast a priori information available for cardiovascular diseases and pave the way of reducing the need for expensive to collect and manually label electrocardiograms.  

\bibliographystyle{IEEEtran} 
\bibliography{mybib}

\onecolumn
\section*{\large Supporting information:}
\section*{\large Training neural networks with synthetic electrocardiograms}
\vspace{2cm}

\renewcommand{\thefigure}{SI \arabic{figure}}
\setcounter{figure}{0}

\renewcommand{\thetable}{SI \Roman{table}}
\setcounter{table}{0}

\setcounter{page}{1}

\begin{figure}[h]
\centering
\includegraphics[width=0.8\textwidth]{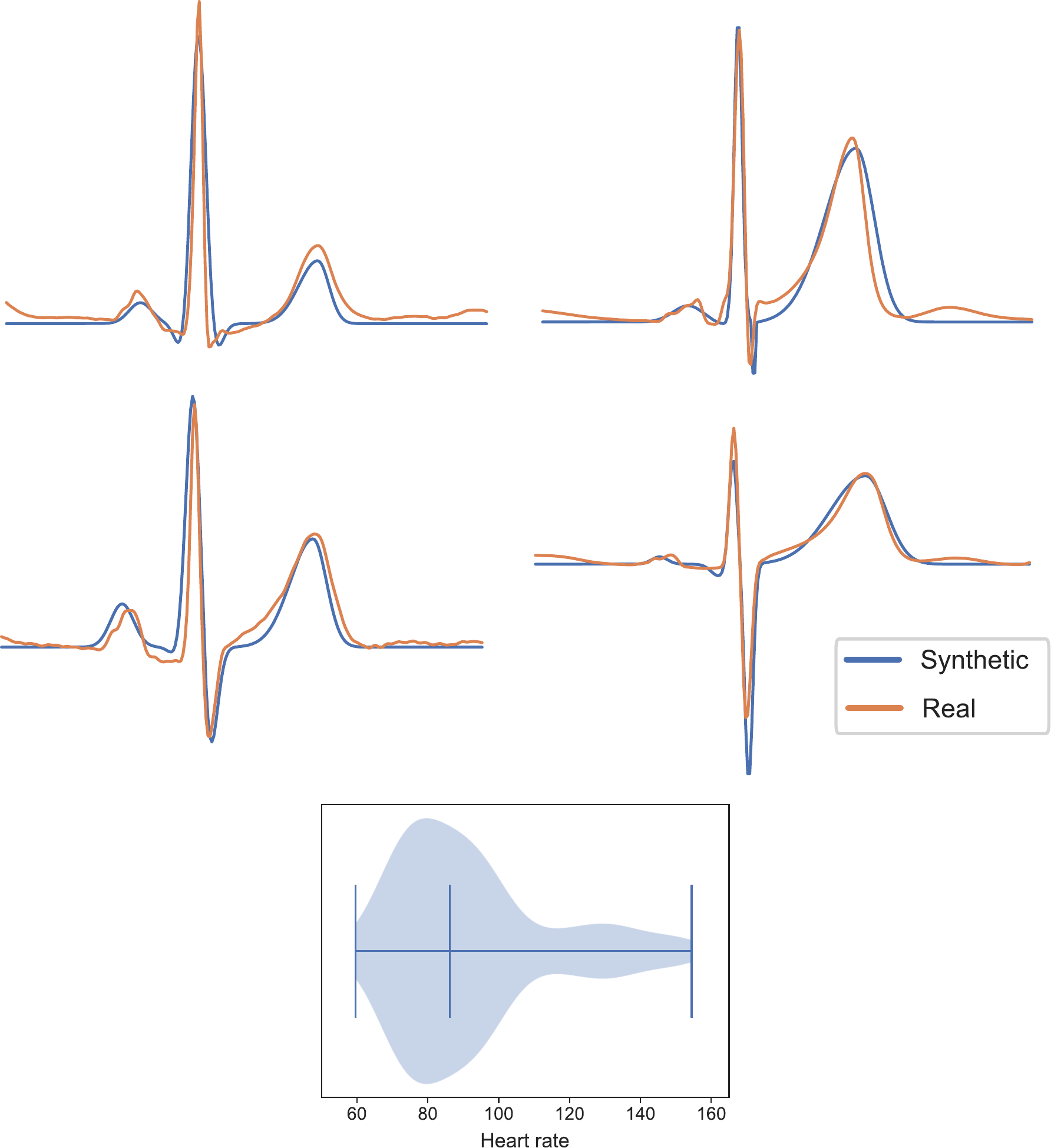}
\caption{Examples of the model fitted to real ECG waveforms that are used to validate the used model input parameters and the heart rate distribution of the ECG-GUDB database.}
\label{fig:waveform_fittings}
\end{figure}

\begin{table}[h]
\centering
\caption{ Parameters for the synthetic signal generator. Single values indicate constancy in random signal generations and values in square brackets are the lower and upper uniform distribution limit. These ranges are scaled by C and listed limits correspond a case C = 1.}
\label{tab:initial ranges}
    \begin{tabular}{lrrrrr}
    \toprule
    Waveform      & p  & q   & r    & s   & t   \\
    \\
    a (amplitude) & [0.05, 0.2] & [-0.05, -0.2] & [0.8, 1.2] & [-0.05, -0.2] & [0.1, 0.6]  \\
    b (width)     & [0.065, 0.085] & [0.03, 0.08] & [0.06, 0.085] & [0.03,0.08]  & [0.085,0.21]  \\
    d (delay to r-peak) & [-0.12, -0.18]  & [-0.03, -0.05]  & 0 & [0.03, 0.05]  & [0.2, 0.25]  \\
    m (asymmetry)       &  1     & 1      & 1     & 1      & [1,3]  \\
    \midrule
    RR-interval         & $\mu$  & $f_b$ & $\beta$&        &        \\
                        & [0.75, 1.0]  & 0.28  & 0.1 &            \\
    \midrule
    Noise               &$\sigma$& $\alpha$& $\rho$&       &        \\
                        & [0, 0.17e-3]  & [0, 0.67]  & [0, 4e-3] &        &        \\
    \bottomrule
    \multicolumn{6}{l}{* $\rho$ during scaling is multiplied by $\alpha^2$ to decrease noise power when $\alpha$ is small.}\\
    \multicolumn{6}{l}{* limits of r are not scaled unless stated otherwise.}  \\
     
    \end{tabular}

\end{table}

\begin{table*}[h]
\centering
\caption{Summary of model performances. }
\label{tab:result-summary}
\begin{tabular}{lrrrrr}
\toprule
                Model &  Loss train &  ROC-AUC train &  Loss test &  ROC-AUC test &  F1-score \\
\midrule
  w/o artefacts C = 0 &       0.005 &          1.000 &      0.414 &         0.921 &     0.677 \\
w/o artefacts C = 0.5 &       0.007 &          1.000 &      0.338 &         0.933 &     0.747 \\
  w/o artefacts C = 1 &       0.011 &          0.999 &      0.263 &         0.927 &     0.764 \\
w/o artefacts C = 1.5 &       0.023 &          0.997 &      0.166 &         0.948 &     0.817 \\
  w/o artefacts C = 2 &       0.031 &          0.994 &      0.106 &         0.979 &     0.919 \\
  w/o artefacts C = 3 &       0.066 &          0.976 &      0.069 &         0.985 &     0.972 \\
\midrule
   w/ artefacts C = 0 &       0.013 &          0.997 &      0.184 &         0.955 &     0.856 \\
 w/ artefacts C = 0.5 &       0.016 &          0.998 &      0.128 &         0.958 &     0.872 \\
   w/ artefacts C = 1 &       0.024 &          0.996 &      0.103 &         0.970 &     0.934 \\
 w/ artefacts C = 1.5 &       0.038 &          0.990 &      0.103 &         0.972 &     0.941 \\
   w/ artefacts C = 2 &       0.050 &          0.983 &      0.091 &         0.987 &     0.959 \\
   w/ artefacts C = 3 &       0.085 &          0.957 &      0.085 &         0.985 &     0.973 \\
\midrule
      Fiducial points &       0.004 &          1.000 &      0.413 &         0.939 &     0.731 \\
          RR interval &       0.006 &          1.000 &      0.371 &         0.947 &     0.738 \\
       Waveform shape &       0.008 &          1.000 &      0.111 &         0.952 &     0.854 \\
                Noise &       0.052 &          0.987 &      0.089 &         0.974 &     0.933 \\
\midrule
   Randomized r C = 1 &       0.064 &          0.978 &      0.083 &         0.973 &     0.882 \\
   Randomized r C = 2 &       0.064 &          0.979 &      0.088 &         0.950 &     0.796 \\
   Randomized r C = 3 &       0.060 &          0.983 &      0.104 &         0.944 &     0.738 \\
\midrule
         Only r C = 3 &       0.060 &          0.976 &      0.118 &         0.975 &     0.868 \\
       Only qrs C = 3 &       0.057 &          0.980 &      0.122 &         0.977 &     0.885 \\
       Only r,t C = 3 &       0.067 &          0.976 &      0.077 &         0.984 &     0.981 \\
\midrule
  Real w/o artefacts  &       0.010 &          0.998 &      0.113 &         0.976 &     0.936 \\
    Real w/ artefacts &       0.018 &          0.995 &      0.122 &         0.982 &     0.958 \\
\bottomrule
\end{tabular}
\end{table*}

\begin{figure*}
\centering
\includegraphics[width=0.8\textwidth]{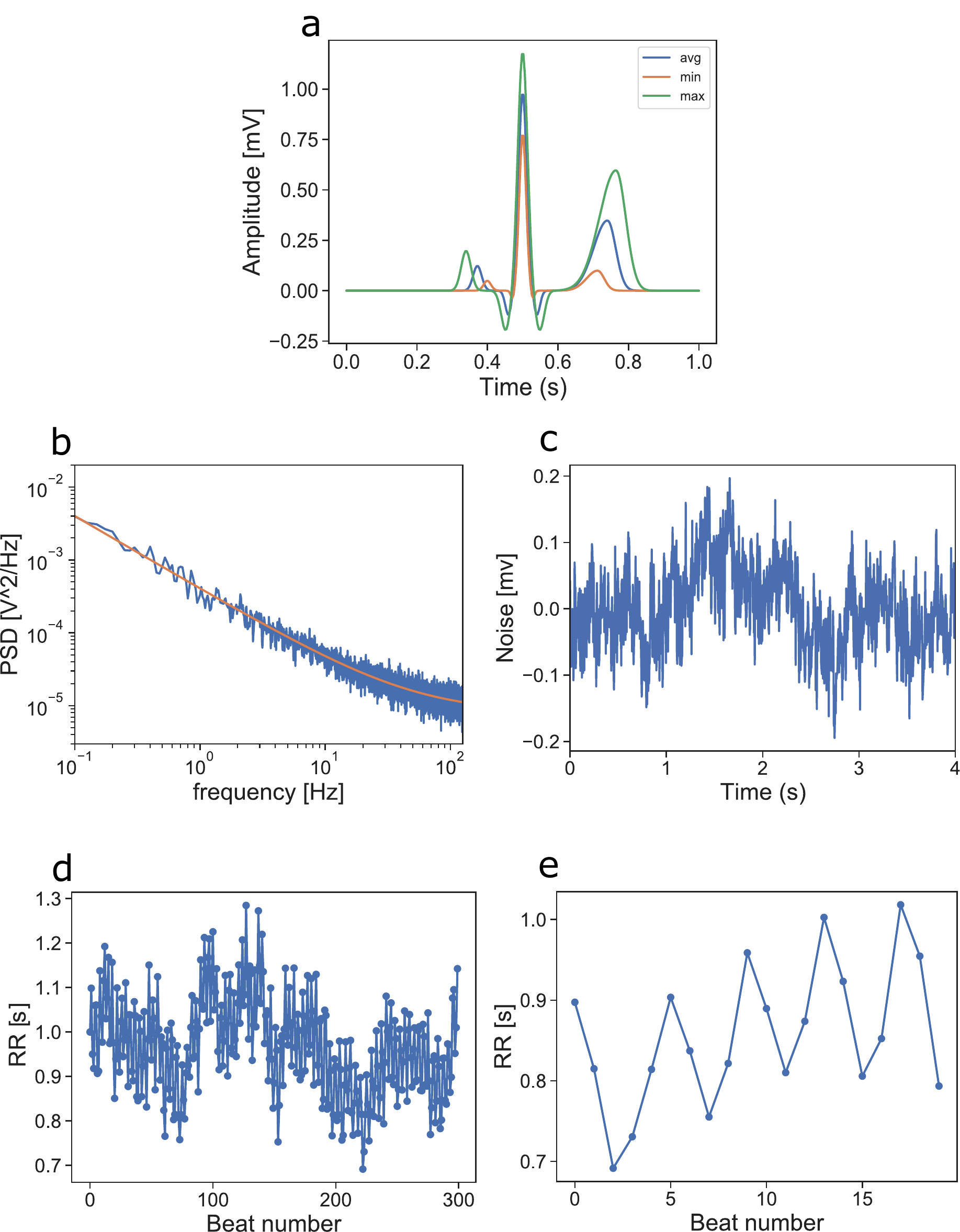}
\caption{a) Synthetic waveforms generated with minimum, maximum and average values when C = 1. This roughly covers the physiological variation in a normal ECG. Any parameter combination is possible e.g. tall and narrow.  b) Synthetic noise generation from a power spectral density (PDF) with 1/f noise and white noise. The flattening part at higher frequencies shows white noise being more prominent where as in the lower frequencies the 1/f noise is stronger. The orange smooth line is the modeled input PSD from which the time domain noise realization in c) is generated. The blue trace in b) is the PSD computed from the generated time domain noise. d) Synthetic RR-intervals generated with $\mu$ = 1 s, $f_b$ = 0.28 Hz and typical values in the stochastic component that includes long-term beat interval correlations\cite{kantelhardt2003modeling}. A 300 beat long stream of RR-intervals is shown and in e) a zoomed in version of it where breathing modulation is apparent.}
\label{fig:noise_model} 
\end{figure*}

\begin{figure*}
\centering
\includegraphics[width=\textwidth]{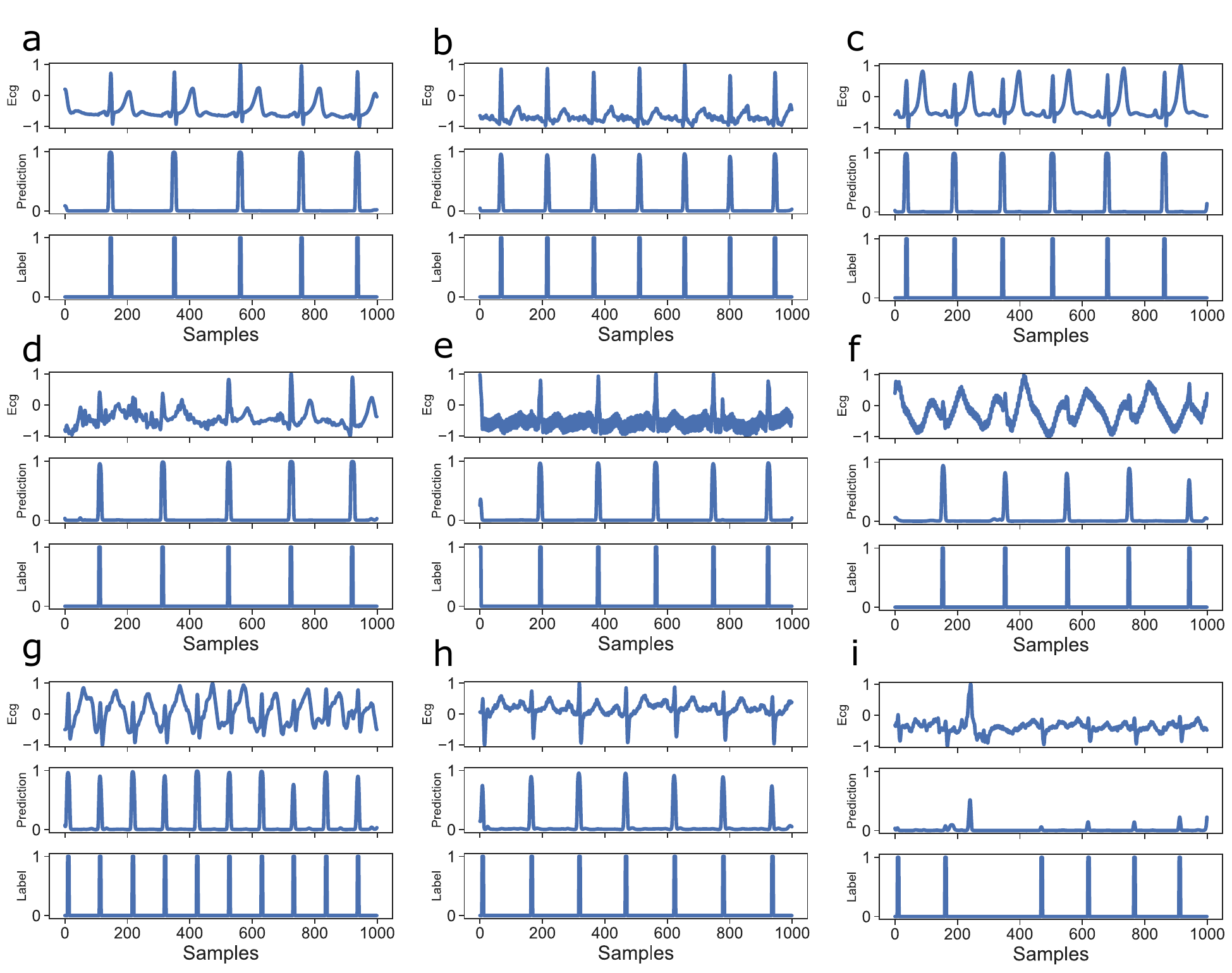}
\caption{ Examples of the model (C = 3 w/o artefacts) predictions with corresponding labels and signals: a) and b) typical high quality ECGs, c) abnormally high t-wave, d) partly noisy segment, e) recording with significant mains interference, f) high baseline wander and mains interference, g) high heart rate and abnormally high t-wave, h) steep  downward deflection of s-wave, i) artefact spike causing unreliable prediction. Such spikes have lower impact when overlapping segments are used.}
\label{fig:predictions}
\end{figure*}


\end{document}